\documentclass[sigconf,nonacm]{acmart}

\setcopyright{none}
\renewcommand\footnotetextcopyrightpermission[1]{}
\usepackage{graphicx}
\usepackage{booktabs}
\usepackage{balance}
\usepackage{stfloats}  
\usepackage{url}       
\usepackage[hyphens]{xurl} 
\usepackage{placeins}  
\usepackage{microtype} 

\newcommand{\code}[1]{\mbox{\texttt{#1}}}

\title{A Framework for Low-Latency, LLM-driven Multimodal Interaction on the Pepper Robot}

\author{Erich Studerus}
\email{erich.studerus@fhnw.ch}
\orcid{0000-0003-4233-0182}
\affiliation{%
  \institution{Institute for Information Systems}
  \institution{University of Applied Sciences and Arts Northwestern Switzerland}
  \city{Basel}
  \country{Switzerland}
}

\author{Vivienne Jia Zhong}
\email{viviennejia.zhong@fhnw.ch}
\orcid{0000-0003-0605-5291}
\affiliation{%
  \institution{Institute for Information Systems}
  \institution{University of Applied Sciences and Arts Northwestern Switzerland}
  \city{Basel}
  \country{Switzerland}
}

\author{Stephan Vonschallen}
\email{stephan.vonschallen@fhnw.ch}
\orcid{0009-0001-4262-938X}
\affiliation{%
  \institution{Institute for Information Systems}
  \institution{University of Applied Sciences and Arts Northwestern Switzerland}
  \city{Basel}
  \country{Switzerland}
}



\begin{document}

\begin{abstract}
Despite recent advances in integrating Large Language Models (LLMs) into social robotics, two weaknesses persist. First, existing implementations on platforms like Pepper often rely on cascaded Speech-to-Text (STT)$\rightarrow$LLM$\rightarrow$Text-to-Speech (TTS) pipelines, resulting in high latency and the loss of paralinguistic information. Second, most implementations fail to fully leverage the LLM's capabilities for multimodal perception and agentic control. We present an open-source Android framework for the Pepper robot that addresses these limitations through two key innovations. First, we integrate end-to-end Speech-to-Speech (S2S) models to achieve low-latency interaction while preserving paralinguistic cues and enabling adaptive intonation. Second, we implement extensive Function Calling capabilities that elevate the LLM to an agentic planner, orchestrating robot actions (navigation, gaze control, tablet interaction) and integrating diverse multimodal feedback (vision, touch, system state). The framework runs on the robot's tablet but can also be built to run on regular Android smartphones or tablets, decoupling development from robot hardware. This work provides the HRI community with a practical, extensible platform for exploring advanced LLM-driven embodied interaction.
\end{abstract}

\maketitle

\section{Background and Summary}

Recent advancements in Large Language Models (LLMs) have fundamentally transformed Human-Robot Interaction (HRI)~\cite{kim2024survey}, enabling open-domain dialogue that replaces scripted systems with natural, context-aware conversation~\cite{pinto2025designing, zhangllm2023}. Several recent studies have integrated LLMs into social robots like Pepper~\cite{banna2025beyond, becerra2024improving, bertacchini2023social, billing2023language, chen2024chatgpt, lim2023sign, pinto2025designing, sievers2025retrieving}. While early implementations primarily used LLMs for dialogue generation, more recent work explores their potential for task planning, gesture generation, and multimodal behavior control~\cite{raptis2025agentic, salimpour2025embodied}.

However, two major barriers prevent these implementations from fully realizing the LLM's potential for multimodal, embodied interaction. First, most existing implementations rely on cascaded pipelines that sequentially process speech through separate Speech-to-Text (STT), LLM response generation, and Text-to-Speech (TTS) stages. This architecture introduces two critical problems: Each processing stage adds incremental delays, resulting in accumulated latency that prevents fluid real-time dialogue~\cite{du2025mtalk, sarim2025direct}. Consequently, recent LLM implementations on platforms like the Pepper robot report system response times ranging from 3.84 to 8.96 seconds~\cite{hanschmann2024designing, pinto2025designing}, far exceeding the 1--2 second threshold established as acceptable for natural conversation flow~\cite{inoue2025noise, pelikan2023managing, shiwa2008quickly, wang2025effects}. Additionally, the conversion of audio to text and back to audio discards paralinguistic information, such as prosody, intonation, and emotional cues, that are crucial for natural, empathetic human-robot interaction~\cite{du2025mtalk, pinto2025designing}.

Second, despite recent progress, most implementations still fail to fully leverage the LLM's capabilities for multimodal perception and agentic control. Achieving embodied AI requires the LLM to convert abstract natural language instructions into concrete, executable actions and integrate diverse sensor feedback in real-time~\cite{abbo2025blind, janssens2025multimodal, salimpour2025embodied}. This requires mechanisms like Function Calling (also known as ``Tool Use'')~\cite{becerra2024improving, raptis2025agentic, salimpour2025embodied}, where the LLM autonomously selects and invokes predefined functions with appropriate parameters. Through Function Calling, the LLM can orchestrate robot actions (navigation, gaze control, tablet interaction) and actively direct its perception by triggering targeted sensor data acquisition (e.g., camera images)~\cite{abbo2025blind, shi2025hribench}. Additionally, the system must integrate asynchronous, event-driven inputs from haptic sensors~\cite{tsirka2025touch} and device interfaces that provide contextual information about physical interactions and system state. The LLM's native multimodal capabilities then process these diverse data streams to achieve embodied grounding in the physical environment~\cite{openai2024realtime, openai2025gpt}.

\begin{figure}[!t]
  \centering
  \includegraphics[width=\columnwidth]{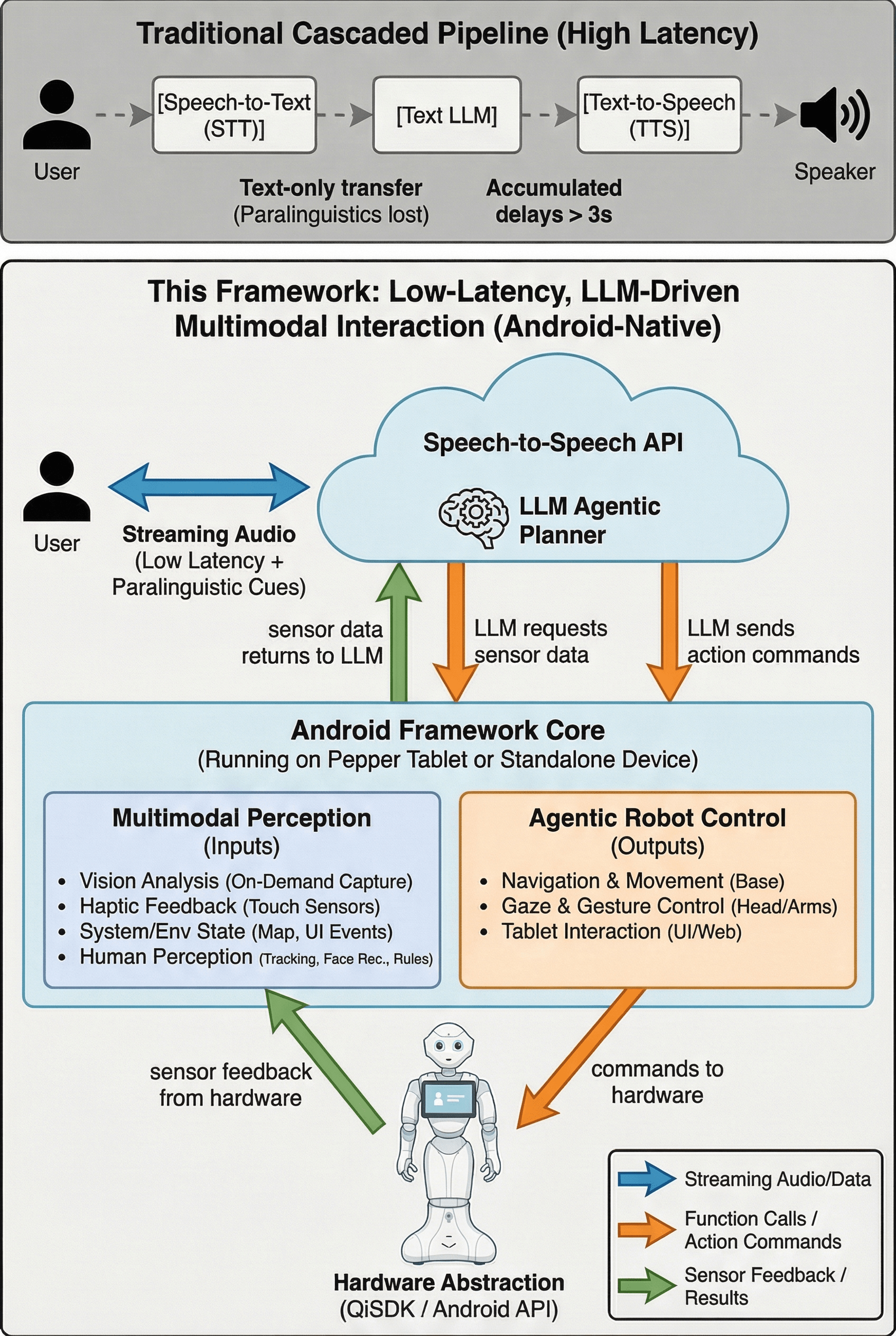}
  \caption{System architecture comparison showing traditional cascaded pipelines versus the proposed low-latency, multimodal framework.}
  \Description{A diagram comparing traditional STT-LLM-TTS pipelines with the proposed framework using end-to-end S2S models.}
  \label{fig:architecture}
\end{figure}

This submission presents an open-source Android framework for the Pepper robot that overcomes these barriers through two key innovations (see Figure~\ref{fig:architecture}): (1) Integration of end-to-end S2S models~\cite{openai2024realtime, openai2025gpt, gemini2025live, fu2025personalized} to achieve low-latency streaming interaction while preserving paralinguistic cues, and (2) extensive Function Calling capabilities that enable the LLM to orchestrate robot actions (navigation, gaze control, tablet interaction) and integrate diverse multimodal feedback (vision analysis, haptic feedback, system state). The framework provides a modular, easily deployable reference implementation that facilitates wider adoption of modern LLM features in embodied platforms. Its design is informed by insights from a prior field study investigating communication dynamics between older adults and an LLM-powered robot~\cite{zhong2025integrating}.

\section{Purpose}

The primary purpose of this framework is to provide the HRI community with a reusable, open-source framework that addresses the latency and limited agentic control challenges outlined above. It establishes a practical blueprint for using an LLM as an agentic orchestrator for multimodal robot behavior---an architectural approach that can be adapted to other robotic platforms. The framework is particularly valuable given the practical challenges facing Pepper researchers: following Aldebaran's court-ordered liquidation on June 2, 2025~\cite{aldebaran2025}, official support for Pepper has become uncertain, and the platform's Android/Java ecosystem and aging hardware (Android 6.0, API level 23)~\cite{qisdk2019} pose non-trivial integration barriers. Furthermore, a key purpose of this contribution is to simplify deployment for HRI research. The framework is designed to run both on the Pepper robot and on standard Android smartphones or tablets, decoupling the development cycle from physical robot hardware. This enables researchers to develop, test, and prototype interaction logic without requiring access to expensive robotic platforms.

\section{Characteristics}

\subsection{Architectural Advantage}

\textbf{Android-Native Execution:} The entire application runs natively on Android devices---either on the Pepper robot's onboard tablet (controlling the robot's physical body via QiSDK) or on standard Android smartphones and tablets. This self-contained architecture eliminates the need for external computers or bridging software common in previous LLM-HRI integrations~\cite{banna2025beyond, bertacchini2023social, billing2023language}, simplifying deployment and use.

\textbf{Dual Build-Flavor System:} The framework can be deployed either on the Pepper robot or on standard Android smartphones and tablets through two build configurations. Robot-specific functions adapt automatically: The \code{analyze\_vision} tool, for instance, accesses the robot's camera system (via QiSDK) when running on Pepper, but uses the device's front camera when running on a smartphone or tablet. Navigation and movement commands control the physical robot or are simulated accordingly. All other tools, including internet search, weather queries, games, and conversational features, operate identically regardless of the deployment platform.

\subsection{Core Interaction Engine: Speech-to-Speech Models}

The framework's low-latency capabilities are powered by end-to-end S2S models from multiple providers (OpenAI~\cite{openai2024realtime, openai2025gpt}, Azure OpenAI, x.ai~\cite{grok2025}, Google~\cite{gemini2025live}). A central \code{TurnManager} orchestrates the robot's state (listening, thinking, speaking). Key advantages include:

\textbf{End-to-End Speech-to-Speech (S2S) Model:} By processing audio directly, the system bypasses the cumulative latency of separate STT/TTS stages. Furthermore, unlike cascaded systems, where information is lost during text conversion, the S2S model can process paralinguistic cues like user intonation. Consequently, its generated speech is not merely text being read aloud; the model adapts its own intonation, tone, and pacing to match the content. While traditional TTS pipelines can also achieve prosody control through explicit SSML markup, this requires complex prompt engineering and rule definitions. The S2S model, on the other hand, generates contextually appropriate prosody intrinsically.

\textbf{Duplex Communication:} While S2S APIs are designed to process user input while simultaneously streaming back responses, our implementation adapts to hardware constraints. Due to the lack of echo cancellation on the Pepper robot, the microphone is intentionally muted during speech output. To compensate, a key usability feature is the ability for the user to tap the status capsule (the interactive state indicator at the bottom of the screen) to instantly interrupt the robot's speech, ensuring a user-controlled conversational experience. The system's duplex nature is still leveraged to process asynchronous, non-auditory inputs (e.g., touch sensor data, tool results) during speech output.

\textbf{Advanced Language and Voice Flexibility:} The models support multiple built-in voices and can respond in a wide range of languages with on-the-fly language switching; a user can change their spoken language mid-conversation, and the model will understand and adapt without requiring configuration changes.

\subsection{LLM as an Agentic Planner: Tool Library}

\begin{figure}[!t]
  \centering
  \includegraphics[width=0.95\columnwidth]{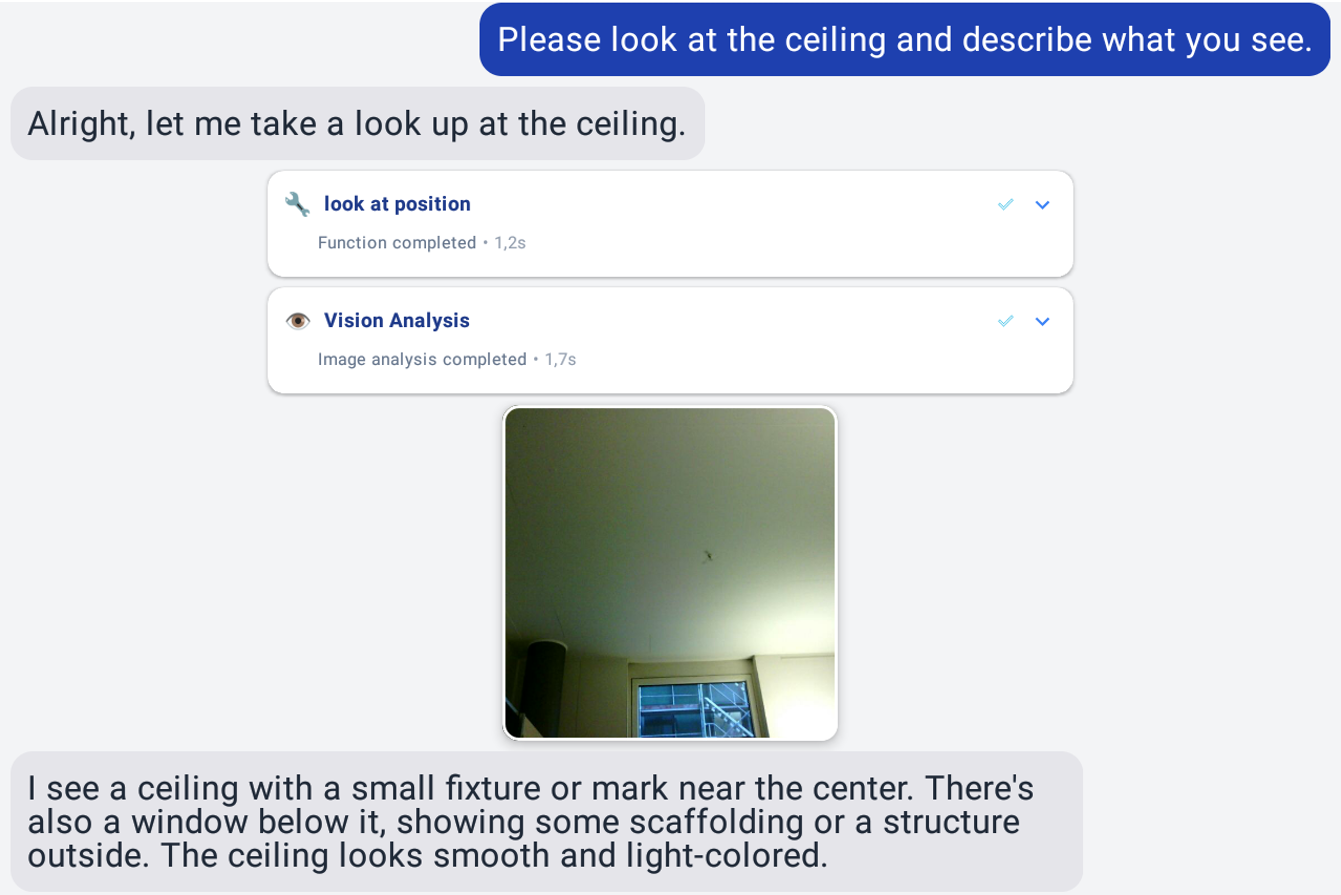}
  \caption{User query triggers multi-step tool sequencing with expandable UI cards showing execution details.}
  \Description{Screenshot showing the chat interface with expandable function call cards.}
  \label{fig:toolcards}
\end{figure}

The framework elevates the LLM to an agentic planner through a rich library of function calls. The LLM can autonomously sequence these tools to fulfill complex requests. For instance, the query ``What do you see at the ceiling?'' triggers a multi-step sequence (see Figure~\ref{fig:toolcards}): First, the LLM invokes \code{look\_at\_position} with calculated 3D coordinates to direct the robot's gaze upward. Then, it calls \code{analyze\_vision}, which captures an image with the robot's front camera and sends it to the S2S backend. The model's native multimodal capabilities process the image directly, enabling it to describe what it sees. Available tools include navigation (movement, floor mapping), multimodal interaction (vision analysis, gaze control), information retrieval (web search, weather, date/time), and interactive entertainment (quizzes, tic-tac-toe, memory game, jokes, drawing game, melody playback, YouTube videos). All tool executions are rendered in the User Interface (UI) as expandable function cards for full transparency.

\FloatBarrier
\subsection{Multimodal Perception and Context}

The framework integrates various sensory inputs to provide the LLM with contextual awareness of its environment.

\textbf{Visual Perception:} Rather than continuous video streaming, the system uses an event-driven architecture where the \code{analyze\_vision} function captures images on demand---triggered by user requests, the model's reasoning, or physical events like obstacle detection. The function sends a camera image as input to the S2S model, which natively supports multimodal input, enabling direct analysis of visual data without requiring separate vision APIs.

\textbf{Haptic Perception:} The robot responds to physical interaction through its various touch sensors on its head, hands, and bumpers. A touch event sends a contextual message (e.g., \texttt{[User touched my right hand]}) directly to the LLM, allowing it to generate a socially appropriate and context-aware response to physical contact.

\textbf{Human Perception:} A custom head-based perception system runs locally on Pepper's head computer, providing stable person tracking and face recognition via WebSocket streaming. This replaces the unreliable QiSDK PeoplePerception metadata with a dedicated YuNet/SFace-based solution~\cite{wu2023yunet} that processes biometric data entirely on-device. An event-based rule system allows researchers to define custom rules that automatically trigger AI responses based on perception events (e.g., person recognized, person approached within interaction distance). Rules can be configured with conditions, cooldowns, and different action types (e.g., trigger an immediate response, update the context of the LLM without requesting a response).

\textbf{System and Environmental State:} The LLM is grounded in its physical context by receiving information about its state and environment. This includes map locations, game interactions, hardware status (e.g., an open charging flap), and error or success messages from tool calls. For example, a movement blocked by an obstacle automatically triggers an \code{analyze\_vision} call, allowing the LLM to reason about the cause of the failure.

\subsection{High Configurability and User Control}

An in-app settings panel allows for runtime adjustments of the entire interaction pipeline. Users can switch between S2S models across four providers: OpenAI (gpt-realtime, gpt-realtime-mini, gpt-4o-realtime-preview, gpt-4o-mini-realtime-preview), Azure OpenAI, x.ai (Grok Voice Agent API~\cite{grok2025}), and Google (Gemini Live API~\cite{gemini2025live}). Azure OpenAI additionally provides network-level isolation and customer-managed encryption keys~\cite{azure2024privacy} for enterprise compliance requirements. Users can also choose between two audio input modes: Direct Audio or STT (speech is first transcribed to text via Azure Speech Services and then sent to the S2S backend). The latter can perform better for some regional dialects, such as Swiss German, and provides confidence scores~\cite{azure2025speech} that inform the LLM when transcription quality may be uncertain.

\subsection{Dependencies, Installation, and Usage}

The framework is built on a modern Android toolchain (API level~35, Java~17) and notably works without the deprecated Pepper SDK plugin, ensuring compatibility with the latest Android Studio versions and future-proofing the codebase. Key dependencies include the SoftBank Robotics QiSDK and OkHttp, all managed via Gradle.

The setup process is straightforward and requires minimal configuration. Only a single API key from one of the supported providers is needed to enable the core conversational features. All other API keys are optional and enable supplementary services: Tavily for internet search, OpenWeatherMap for weather information, Azure Speech for alternative speech recognition, and YouTube Data API for video playback. A comprehensive guide in the README.md details the setup process.

\subsection{Availability}

The framework is provided as an open-source artifact under the MIT License at \url{https://github.com/studerus/pepper-android-realtime-chat}. The repository includes comprehensive documentation and setup instructions.

\section{Code/Software}

\subsection{Code Structure and Key Components}

The software is an Android application written entirely in Kotlin, built with Gradle, and architected around a build flavor system (as described in Section~3.1) to separate hardware-dependent code from the core application logic. The codebase leverages modern Kotlin features including coroutines for structured concurrency, data classes, and Hilt for dependency injection. This results in a highly modular and extensible structure divided into three primary source sets:

\begin{enumerate}
    \item \code{app/src/main/} (Shared Core): Contains most code shared between \texttt{pepper} and \texttt{standalone} flavors. Key components include \code{ChatActivity.kt} (UI, lifecycle, and service orchestration), \code{TurnManager.kt} (conversational state machine for listening, thinking, speaking), \code{RealtimeSessionManager.kt} (WebSocket communication with S2S backends), \code{ToolRegistry.kt} (agentic system core for registering, validating, and executing function calls), and the \texttt{robot/} sub-package with abstraction interfaces (\code{RobotController}, \code{RobotLifecycleBridge}) that define the contract for robot interaction, allowing the main module to remain hardware-agnostic.
    
    \item \code{app/src/pepper/} (Pepper Flavor): Contains Pepper-specific implementations using QiSDK. For example, \code{MovementController.kt} translates abstract movement commands into QiSDK actions.
    
    \item \code{app/src/standalone/} (Standalone Flavor): Provides implementations for standard Android devices---either stubs (e.g., \code{MovementController.kt} logs actions) or adapted implementations (e.g., \code{VisionService.kt} uses Camera2 API).
\end{enumerate}

\subsection{Configuration and Data Formats}

Build configuration is managed via \code{app/build.gradle}, which injects API keys from \code{local.properties} (excluded from version control) into \texttt{BuildConfig} at compile time. A template file (\code{local.properties.example}) guides key setup.

\subsection{Code Maintenance and Supplemental Documentation}

Contributions via GitHub Issues for bug reports and feature requests, as well as Pull Requests for enhancements, are welcome and will be reviewed. The primary source of documentation is the comprehensive README.md file, which provides step-by-step installation instructions, detailed feature explanations, API key setup guidance, and troubleshooting tips. Additional documentation includes inline code comments and architectural diagrams.

\section{Usage Notes}

\subsection{Utility and Extensibility for HRI Research}

The framework is designed for extensibility and adaptation. Researchers can extend its capabilities by implementing new tools that conform to the Tool interface, modify robot personalities through the system prompt, or integrate additional sensor modalities. While implemented on Pepper's Android platform (QiSDK), the architectural principles, particularly the abstraction layer separating core logic from hardware-specific implementations, are transferable to other Android-based humanoid robots. Beyond Android, the conceptual approach (S2S model integration, Function Calling for robot control, multimodal grounding) provides a blueprint adaptable to other robotic platforms. For advanced use cases on Pepper, the framework establishes an SSH connection at startup, enabling access to low-level NAOqi functions not exposed by QiSDK. This connection is also used to launch the head-based people perception system on the robot's head computer.

\subsection{Ethical Considerations and Responsible Use}

Our implementation may expose researchers to data security issues. In particular, the current method of storing API keys in \texttt{BuildConfig} is suitable for development but is not secure for production deployments. Furthermore, user data, including audio and images, is transmitted to third-party cloud services like OpenAI and Azure when their respective features are active. A detailed summary of these considerations, including instructions on how to disable specific data-transmitting features, is provided in the ``Security \& Privacy'' section of the README.md file.

\subsection{Limitations and Outlook}

Several limitations should be acknowledged: First, Pepper lacks articulated lips, so lip-syncing---relevant for robots like Furhat---is not addressed. Second, while the architectural principles are designed to be transferable, the current implementation is validated only on the Pepper platform. Third, the reliance on cloud-based APIs introduces dependency on external services and potential latency variations based on network conditions. Finally, developers using speech-to-speech models have reduced control over specific phrasings compared to text-based pipelines.

Despite these limitations, this framework represents a significant step towards fluid, low-latency, and agentic human-robot interaction. By open-sourcing this tool, we aim to accelerate the adoption of multimodal, agentic LLM capabilities in social robotics, encouraging the community to extend these principles to new platforms and interaction scenarios.

\balance
\bibliographystyle{ACM-Reference-Format}
\bibliography{references}

\end{document}